\documentclass[11pt]{article}
\usepackage{geometry}
\usepackage{coling2020}
\usepackage{times}
\usepackage{url}
\usepackage{latexsym}
\usepackage{microtype}
\usepackage{tikz}
\usepackage{subcaption}
\usepackage{caption}

\usetikzlibrary{shapes,arrows}
\usepackage{amsmath,bm,times}

\colingfinalcopy 

\title{QiaoNing at SemEval-2020 Task 4: Commonsense Validation and Explanation system based on ensemble of language model}

\author{Pai Liu \\
  Northeastern University, Shenyang, China \\
  {\tt pai.liu.1998@gmail.com} \\}

\date{}

\begin{document}
\maketitle
\begin{abstract}
 In this paper, we present language model system submitted to SemEval-2020 Task 4 competition: "Commonsense Validation and Explanation". We participate in two subtasks for subtask A: validation and subtask B: Explanation. We implemented with transfer learning using pretrained language models (BERT, XLNet, RoBERTa, and ALBERT) and fine-tune them on this task. Then we compared their characteristics in this task to help future researchers understand and use these models more properly. The ensembled model better solves this problem, making the model's accuracy reached 95.9\% on subtask A, which just worse than human's by only 3\% accuracy. 
\end{abstract}

\section{Introduction}

\blfootnote{
    %
    %
    \hspace{-0.65cm} 
    This work is licensed under a Creative Commons Attribution 4.0 International License. License details: \url{http://creativecommons.org/licenses/by/4.0/}.
    %
    %
    %
    %
}

This task \cite{wang-etal-2019-make}is to evaluate how well a model can do for sense making on English data set by its ability to judge whether natural language sentences are not in line with common sense and explain the reasons. 
To thoroughly evaluate the model, three subtasks are designed in \cite{wang-etal-2020-semeval}:
a) Subtask A Validation - Given two statements with similar structures, the task is to discern which statement relatively makes sense and which one does not.
b) Subtask B Explanation(Multiple-choices) - From the three candidate reasons given, choose the reason that is most likely to explain a statement that doesn't make sense.
c) Subtask C Explanation(Generation) - Generating explanations for non-common sense statements.
Natural Language Understanding (NLU) has drawn an increasing amount of research attention recently. Although some well-performed end-to-end model even show better performances than humans on some benchmarks, they are still far from human on common sense. If a model lack of common sense processes data that requires common sense, the performance of the model may drop dramatically, which is a sign for poor robustness. However, building models that are robust to adversarial attacks is essential\cite{DBLP:journals/corr/abs-2003-00120}.
Therefore, the importance of commonsense reasoning is not only confined to NLU system, it should also be expanded to other systems. 


Many similar tasks have been studied and solved before this task was proposed. Previous attempts on solving common sense challenge usually involve heavy utilization of annotated knowledge bases\cite{DBLP:conf/naacl/PengKR15},
rule-based reasoning\cite{DBLP:conf/aaaiss/BaileyHLLM15}, 
or hand-crafted features\cite{DBLP:conf/kr/Schuller14}. 
Over time, the advantages of the knowledge base gradually emerged from other methods, and researchers are more inclined to use this method to solve such problems. Simple models based on distributional representations perform poorly on this task, despite doing well on selection preference, but injecting manually elicited knowledge about entity properties provides a substantial performance boost\cite{DBLP:conf/naacl/WangDE18}. 
However, using  annotated knowledge bases is a expensive approach while unsupervised training is done on text corpora which can be cheaply curated.
Nowadays, neural LMs having achieved great success being used as feature representations for a sentence, or a paragraph, which improves NLP applications in a large scale such as question answering, sentiment analysis, machine translation, etc. The combined evidence suggests that LMs trained on a extensive amount of unlabeled data can capture many aspects of natural language and the world’s knowledge, especially commonsense information. The success of previous work  of using language models to solve Winograd Schemas Challenge further confirms the inferences above.
By looking at probability ratio at every word position. 
Trieu\shortcite{DBLP:journals/corr/abs-1806-02847} used LM to score multiple choice questions posed by commonsense reasoning tests. This work inspired us to make use of the capability of capturing commonsense information. Whereas unsupervised approach doesn't need expensive knowledge bases and success of previous LM models, we present a transformer-based approach where a pretrained language model has been fine-tuned on data set of this task. We voted and weighted the results of different models and the same model with different training parameters. Finally, the ensembled model using the weighting method works best.
%


It is found that the results of some models are similar, while others are different. This is related to the structure of different models, the training data and training time, etc. Therefore, we believe the performance can be improved by ensemble learning for only complementary models can correct each other to get better results. The ensembled model achieves great result, ranking in official test evaluation 6th and 11th place in subtasks A and B, respectively. Our best model achieved in official test evaluation accuracy of 95.9\% for subtask A and 90.8\% for subtask B. In addition, we also made a comprehensive analysis of the data of task A. We found that data can be classified by the structure of the sentence since the data set of task A can be divided into 3 categories, and the details will be explained in Section 3 later. Our model does not deal with different types of data with different approaches separately, but the future model proposed for these characteristics may better solve the task.

\section{Models}
We trained a large variety of different models
and combined the best of them in ensembles\footnote{The parameters of each model are released on \url{https://github.com/zellford/QiaoNing-at-SemEval-2020-Task-4-parameters}} . For every LM model, we just add a feed forward layer and a softmax layer to the end of it without change its original structure. Voting weights and evaluation scores of different models are related. The weight sum equation is defined as:
\begin{equation}
y=\sum_{i=1}^{N} w_{i} p_{i} 
\label{eq:weight_sum}
\end{equation}
where $\mathbf{y}$ is the final prediction, $\mathbf{N}$ is the number of total model in ensembles, $\mathbf{w_{i}}$ is the weight of i-th model and $\mathbf{p_{i}}$ is the prediction of i-th model. In this work, we denote the number of layers (i.e., Transformer blocks) as L, the hidden size as H, the vocabulary embedding size as E,and the number of self-attention heads as A, which is the same definition as the ALBERT. The architecture of the models ensemble is shown as Figure \ref{fig:my_label}.


\subsection{BERT}
With the release of Bidirectional Encoder Representation of Transformers\cite{DBLP:conf/naacl/DevlinCLT19}, the pretrained and fine-tuning approach to many problems become popular. Many similar pretrained models were designed and refresh the score of many benchmark. So we first fine-tuned BERT with our dataset. BERT’s model architecture is a multi-layer bidirectional Transformer encoder based on the original implementation described in \cite{DBLP:conf/nips/VaswaniSPUJGKP17}. 
6 checkpoints with different configuration are provided, whose model sizes and the way to preprocess words are varied. We experimented on $BERT_{Base}$
and $BERT_{Large}$.
\subsection{XLNet}
XLNet\cite{DBLP:conf/nips/YangDYCSL19} has more training corpus than BERT and presents a new pre-trained way that differ from BERT, which can help us discover analyse the effect of data size and pre-training method on the task. XLNet is a language model, which is in line with ELMO, GPT and BERT. However, they are exactly different in some way. To be precise, BERT belongs to autoencoding(AE) language model, which means encoding sequence $\mathbf{x=(x_{1}...x_{T})}$ to sequence $\mathbf{y=(y_{1}...y_{T})}$, while autoregressive language model predicts xi according to $\mathbf{(x_{1}...x_{i-1})}$. Thanks to the use of both the above and the following information, Bert achieved better results than GPT. However, Bert needs to introduce [mask] tags in the pre-train stage to predict these masked tokens from the context, which leads to two main problems: Bert assumed that different [masks] were independent of each other and ignored the correlation between [masks]; The input to BERT contains artificial symbols like [MASK] that never occur in downstream tasks, which creates a pretrain-finetune discrepancy. Therefore, XLNet used two-stream self-attention as the method of "permutation and combination" of input sequences to put the following information in the front, which gives the one-way model the ability to perceive the following information. In addition, XLNet also integrates the relative positional encoding scheme and the segment recurrence mechanism from Transformer-XL.
\subsection{RoBERTa}
Robustly optimized BERT Pretraining Approach(RoBERTa) is a replication study of BERT pretraining , which carefully evaluates the impact of key hyperparameters and training data size without modifying the structure of BERT. We chose Roberta because it can be used as a good comparison object to compare with BERT for the amount of data and the contribution of more adequate training, and the impact of the different training methods compared to XLNet on common sense judgment. First, RoBERTa\cite{DBLP:journals/corr/abs-1907-11692} dynamically changes mask pattern applied to the train data instead of uses the same mask for each training instance in every epoch. Second, to capture the relationship between two sentences, BERT uses next sentence prediction(NSP) in train stage. Nevertheless RoBERTa abandons NSP and presents a new training way called Full-Sentences, which can input multiple sentences. Third, training with large mini-batches. Previous work in Neural Machine Translation has proved that in this way can improve optimization speed and end-task performance. Finally, XLNet used 10 times more data than Bert, resulting performance did indeed soar again. Of course, it also requires longer training.
\subsection{ALBERT}
According to the previous neural language model, a conclusion can be easily drawn that increasing model size, training data and training time will definitely improve LM's performance. As a new LM,on the contrary, ALBERT\cite{DBLP:journals/corr/abs-1909-11942} used less parameter to reduce the need of memory capacity of GPU/TPU and shared parameters weight, which will improve parameter efficiency, to fasten training speed. As for pre-train task, ALBERT presented sentence order prediction(SOP) to replace NSP. The SOP loss uses the same technique as BERT (two consecutive segments in the same document) as a positive example, and uses the same two consecutive segments (but the order is reversed) as a negative example. In addition to the three main optimization points mentioned above, drop-out layer is removed.
\section{Data}

\pgfdeclarelayer{background}
\pgfdeclarelayer{foreground}
\pgfsetlayers{background,main,foreground}

\tikzstyle{sensor}=[draw, fill=black!20, text width=5em, 
    text centered, minimum height=2.5em]
\tikzstyle{sensor1}=[draw, fill=black!30, text width=5em, 
    text centered, minimum height=2.5em]
\tikzstyle{sensor2}=[draw, fill=black!40, text width=5em, 
    text centered, minimum height=2.5em]
\tikzstyle{ann} = [above, text width=5em]
\tikzstyle{naveqs} = [sensor, text width=6em, fill=gray!60, 
    minimum height=12em, rounded corners]
\def\blockdist{2.3}
\def\edgedist{2.5}

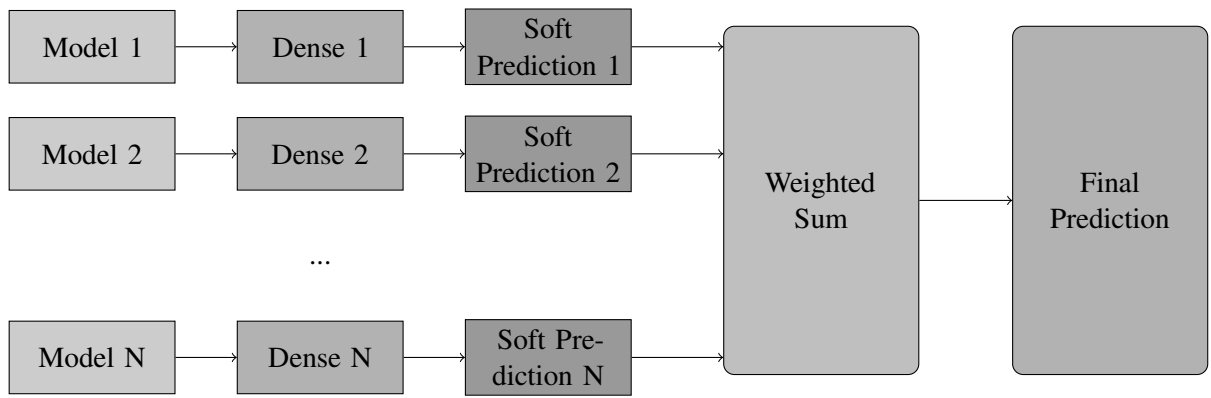
\begin{figure}
 \centering
\begin{tikzpicture}

    \node (naveq)[naveqs, fill=gray!50] {Weighted \\ Sum };
    \path (naveq)+(\blockdist+1.5,0) node (result) [naveqs] {Final \\ Prediction};
    \path (naveq.130)+(-\blockdist-6,0.5) node (item) [sensor] {Model 1};
    \path (naveq.130)+(-\blockdist-3,0.5) node (item_1) [sensor1] {Dense 1};
    \path (naveq.130)+(-\blockdist,0.5) node (item_2) [sensor2] {Soft Prediction 1};
    
    \path (naveq.-140)+(-\blockdist-3,0.25) node (bre){...};
    
    \path (naveq.-140)+(-\blockdist-6,1.7) node (accel) [sensor] {Model 2};
    \path (naveq.-140)+(-\blockdist-3,1.7) node (accel_1) [sensor1] {Dense 2};
    \path (naveq.-140)+(-\blockdist,1.7) node (accel_2) [sensor2] {Soft Prediction 2};
    \path (naveq.-140)+(-\blockdist-6,-1) node (pie) [sensor] {Model N};
    \path (naveq.-140)+(-\blockdist-3,-1) node (pie_1) [sensor1] {Dense N};
    \path (naveq.-140)+(-\blockdist,-1) node (pie_2) [sensor2] {Soft Prediction N};

    \path [draw, ->] (accel) -- (accel_1);
    \path [draw, ->] (accel_1) -- (accel_2); 
    \path [draw, ->] (accel_2) -- (naveq.west |- accel_2);
    
    \path [draw, ->] (item) -- (item_1);
    \path [draw, ->] (item_1) -- (item_2); 
    \path [draw, ->] (item_2) -- (naveq.west |- item_2);
    
    \path [draw, ->] (pie) -- (pie_1); 
    \path [draw, ->] (pie_1) -- (pie_2); 
    \path [draw, ->] (pie_2) -- (naveq.west |- pie_2);
    
    \path [draw, ->] (naveq) -- (result); 
    
    
\end{tikzpicture}
\caption{The architecture of the models ensemble}
\label{fig:my_label}
\end{figure}

Commonsense Validation and Explanation dataset (ComVE) is released by the organizers of this task. Each instance in subtask A consists of 2 similar sentences, but only one of them makes sense while the other does not. As for the subtask B, every invalid sentence are followed by three sentence as the optional explanation, in which only one is correct. For the same invalid sentence in subtask C, 3 referential reasons are given as the correct answer to evaluate the reasons that machine generates. 
\begin{table}[]
    \centering
    \begin{tabular}{|c|c|c|c|}
        \hline
        \bf Type & \bf Choice A & \bf Choice B & \bf Label \\
        \hline
        Sample a) & The sky is blue & The sky is underground & 1 \\
        \hline
        Sample b) & the man fed the snake a mouse & the man fed the mouse a snake & 1\\
        \hline
        Sample c) & The bike overtake the car & The red car went by very fast & 0 \\
        \hline
    \end{tabular}
    \caption{Examples of three types of subtask A test samples}
    \label{tab:my_label_1}
\end{table}
All samples are written by data annotators, then researchers examines them case by case. Ensuring the consistency of samples, three principles are followed: First, make everyone can understand question with their commonsense easily, that is to say complex knowledge is not needed. Second, important words like entities and activities should be contained in confusing reasons in the against-common-sense statements. Third, the reason why a statement is confusing is related to the context. The tasks release the dataset into three different
parts, which are the training dataset, develop dataset and testing dataset. The summary of dataset distribution
is concluded in the Table \ref{tab:my_label_1}.
From the table, it is obviously that organizers intentionally avoid the impact of uneven distribution of answers. 
\subsection{Data Analysis}
At the end of the evaluation, although we tried different hyperparameter and ensemble many well fine-tuned model, the score on developing dataset still became steady. Thus we turn to discover the characteristics of the data. By checking dataset case by case, we find that all samples can be classified into 3 types and examples are shown in Table 1: a) there is only one difference between most wrong samples and correct ones, like numbers, entities, action and etc. b) Keyword order replacement leads to ambiguity in sentence meaning. c) We take the rest as the third category because these examples have no obvious characteristics. Some are talking about the same subject but the sentence expression is completely different, while others are even not talking about the same thing at all. Fortunately, the first two types of examples with clear structure account for most of the dataset. Therefore, processing on certain type confusing samples may further improve the performance. For  a) type sample, masking the different word in sentence pair and calculate the probabilities of original words can be a good method. As for subtask C, when generating the reasons for type b) sample, reason template like "A is ... than B" can be used.
\section{Experiments and Results}
The evaluation metric of first two subtasks is accuracy and we use BLUE to evaluate subtask C. In each experiment, model is trained on 1080Ti, and batch size is set as big as possible to ensure rapid convergence of the model. Due to labels of test dataset are kept by organizers, we tune the hyperparameters on development dataset. Because different language models use different corpora, hyperparameters, and model structures during pre-training, we believe that these differences will affect the performance of the model. In order to more fully reflect the ability of each model on this task, we did not introduce additional training data when fine-tuning, retaining the data provided by the organizer. We only participate in first two subtasks. For subtask A, we input a sentence pair at a time, the output is the label that indicating which sentence is invalid. For subtask B, a multiple choices problem , we concatenate the invalid statement with each optional reason as inputs. Then the scores of each sentence-reason pair are fed into softmax layer to get final output. Finally, we weight each model's result according to evaluation score on developing model, make the result less than 0.5 as 0, otherwise as 1.

In Tables \ref{subtable_a} and \ref{subtable_b}, we report the best results of each model of final test dataset. When fine-tuning different types of BERT, we found that the case-insensitive model is better than the case-sensitive model, and the larger model (BERT large) will perform better. One surprising observation from the table shows the performance of XLNet is far behind the BERT‘s, let alone other improved models based on BERT. We attribute this phenomenon to the way XLNet training, which does not include NSP-like tasks. Although XLNet surpasses BERT on many benchmarks, this sign indicates that it still lacks the ability to understand the relationship between sentences. In contrast, ALBERT's accuracy is among the best. We think it is because ALBERT replaced NSP with SOP as a pre-training task, which forces the model to learn finer-grained distinctions about discourse-level coherence properties. In addition, when we look at some of our ensemble model  predictions that range from 0.4 to 0.6, which means the feasibility of the results is extremely low. The reason for the ambiguous result is that the predictions given by BERT and RoBERTa on these samples are consistent and completely opposite to the predictions of ALBERT. We replaced the ambiguous predictions with predictions of each individual model to find out which model can better solve these samples. So we got the result of the Table \ref{subtable_c}.
   

\begin{table}[htp]
\centering
\begin{minipage}{0.3\textwidth}
\centering

\begin{tabular}[height = 6cm]{l|l}
\hline \bf System & \bf Accuracy \\ \hline
BERT(B) &  83.1  \\
BERT(L) & 90.1  \\
XLNet(B) & 81.6 \\
RoBERTa(B) & 87.4 \\
RoBERTa(L) & 93.5 \\
ALBERT(xxL) & 95.3 \\
 \hline
Ensemble & 95.9 \\
\hline
\end{tabular}
\subcaption{ Subtask A results }
\label{subtable_a}
\end{minipage}
\begin{minipage}{0.3\textwidth}
\centering

\begin{tabular}[height = 6cm]{l|l}
 \hline \bf System & \bf Accuracy \\ \hline
        BERT(L) & 85.7  \\
        XLNet(L) & 90.3 \\
        RoBERTa(L) & 87.0 \\
         \hline
        Ensemble & 91.2 \\
        \hline
\end{tabular}
\subcaption{ Subtask B results }
\label{subtable_b}
\end{minipage}
\begin{minipage}{0.3\textwidth}  
\centering

\begin{tabular}[height = 6cm]{l|l}
\hline \bf System & \bf Accuracy \\ \hline
        BERT(L) & 96.2  \\
        XLNet(B) & 95.7 \\
        RoBERTa(L) & 96.2 \\
        ALBERT(xxL) & 95.3 \\
        \hline
\end{tabular}
\subcaption{ Subtask A post-results }
\label{subtable_c}
\end{minipage}


\centering
\caption{Results on test data where B denotes Base and L denotes Large, and in Table  \ref{subtable_c} ambiguous predictions (between 0.4 and 0.6) are replaced by each model ’s own predictions.}
\end{table}

Obviously, ALBERT has insufficient ability to deal with these problems. Surprisingly, when the overall performance of ALBERT and XLNet differ by 13.7\%, XLNet still has a higher accuracy on this type of problem. Therefore, we dig into the predictions of each model to find the cause of this phenomenon. We found that for two-thirds of the samples our models give exactly the same predictions. These confident samples almost belong to type a), and the correct sentences are often found in the article, on the contrary, the wrong sentences will hardly appear. As the prediction divergence of the model becomes larger, the number of sentences of type b) and type c) gradually increases, but this is not the most obvious change. The most important thing is that the differences between the options of these samples become less obvious. In other words, all the options may not have similar sentences in the corpus. The key to judging these problems is often related to the hidden characteristics of certain words, so this requires our model to understand the comprehensive meaning of each word. Compared with other models, ALBERT has a similar model size, more difficult training tasks and even more data volume, but the performance of the implicit word meaning is not good. It is noteworthy that in order to reduce the amount of parameters, which will be a significant parameter reducing when $H >> E$, ALBERT reduces the dimension of word embedding. Excluding the above-mentioned factors that affect the performance of the model, we think this is because the width of embedding word of ALBERT is at least 6 times smaller than other models that can't fully express the meaning of words.  Finally, the ensembled model achieves better result, ranking in official test evaluation 6th and 11th place in subtasks A and B, respectively.
\section{Conclusion}
The ability of common sense validation and explanation is very important for most models. Most obviously, this will directly affect the rationality of the generated model output. The large amount and diversity of common sense poses great challenges to this task. In addition, many common sense expressions are obscure, thus we need to understand the meaning contained in the vocabulary in order to judge correctly, which further increases the model's requirements for the accuracy of word representation.
The current neural network models are often data-driven, while the annotated data is often limited and requires a lot of manual labeling. In such case, we proposed transfer learning to handle this challenge. From our experiments,
we can draw the following three main conclusions: 
a) Neural language model fully qualified for commonsense validation and explanation. We attribute this to the powerful word and sentence representation capabilities of language models.
b) The inconsistency of task of pre-training and fine-tuning will badly hurt the performance.
c) A larger amount of corpus and more parameters will enhance the common sense of the model. At the same time, the content of the corpus is equally important.

\section*{Acknowledgements}
We thank Yinqiao Li, Chen Xu, Yuxin Wang, Qing Cai and QiaoNing Yang for their insightful discussions. We also thank all anonymous reviewers for their constructive comments.

\bibliographystyle{coling}
\bibliography{semeval2020}

\end{document}